\documentclass[letterpaper, 10 pt, conference]{ieeeconf}  % Comment this line out if you need a4paper

\IEEEoverridecommandlockouts                              % This command is only needed if 
\usepackage{amsthm}
\usepackage{times} % assumes new font selection scheme installed
\usepackage{amsmath} % assumes amsmath package installed
\usepackage{amssymb}  % assumes amsmath package installed
\usepackage{algorithmic}
\usepackage{graphicx}
\usepackage{textcomp}
\usepackage[usenames, dvipsnames, table]{xcolor}
\usepackage{float}
\usepackage[style=ieee,sorting=none,backend=biber, maxbibnames=99]{biblatex}
\usepackage[hidelinks]{hyperref}
\usepackage[ruled,resetcount,vlined]{algorithm2e}
\usepackage{commath}
\usepackage[font=small]{caption}
\usepackage{subcaption}
\usepackage{booktabs}
\usepackage{diagbox}
\definecolor{dusty_blue}{HTML}{56B4E9}
\definecolor{dusty_green}{HTML}{009E73}
\definecolor{dusty_pink}{HTML}{CC79A7}

\colorlet{mylinkcolor}{BrickRed}
\colorlet{mycitecolor}{Green}
\colorlet{myurlcolor}{NavyBlue}

\hypersetup{
  linkcolor  = mylinkcolor,
  citecolor  = mycitecolor,
  urlcolor   = myurlcolor,
  colorlinks = true,
}

\usepackage[capitalize]{cleveref}

\theoremstyle{definition}

\renewcommand{\Cref}[1]{\cref{#1}}
\Crefname{definition}{Def.}{Defs.}

\title{\LARGE \bf
Exploring How Non-Prehensile Manipulation Expands Capability in Robots Experiencing Multi-Joint Failure
}

\author{Gilberto Briscoe-Martinez$^{*}$, Anuj Pasricha, Ava Abderezaei, Santosh Chaganti,\\ Sarath Chandra Vajrala, Sri Kanth Popuri, and Alessandro Roncone% <-this % stops a space
\thanks{$^{*}$ Corresponding author.}
\thanks{GBM is supported by NASA Space Technology Graduate Research
Opportunity Grant 80NSSC22K1211.}% <-this % stops a space
\thanks{All authors are with the Department of Computer Science, University of Colorado Boulder, 1111 Engineering Drive, Boulder, CO USA.
{\tt\small firstname.lastname@colorado.edu}}
}

\begin{document}

\maketitle
\thispagestyle{empty}
\pagestyle{empty}

%%%%%%%%%%%%%%%%%%%%%%%%%%%%%%%%%%%%%%%%%%%%%%%%%%%%%%%%%%%%%%%%%%%%%%%%%%%%%%%%
\begin{abstract}
This work explores non-prehensile manipulation (NPM) and whole-body interaction as strategies for enabling robotic manipulators to conduct manipulation tasks despite experiencing locked multi-joint (LMJ) failures. LMJs are critical system faults where two or more joints become inoperable; they impose constraints on the robot's configuration and control spaces, consequently limiting the capability and reach of a prehensile-only approach. This approach involves three components: i) modeling the failure-constrained workspace of the robot, ii) generating a kinodynamic map of NPM actions within this workspace, and iii) a manipulation action planner that uses a sim-in-the-loop approach to select the best actions to take from the kinodynamic map. The experimental evaluation shows that our approach can increase the failure-constrained reachable area in LMJ cases by 79\%. Further, it demonstrates the ability to complete real-world manipulation with up to 88.9\% success when the end-effector is unusable and up to 100\% success when it is usable.
\end{abstract}

%%%%%%%%%%%%%%%%%%%%%%%%%%%%%%%%%%%%%%%%%%%%%%%%%%%%%%%%%%%%%%%%%%%%%%%%%%%%%%%%
% \input{intro}
\section{Introduction}\label{sec:introduction}

% ``In this world nothing can be said to be certain, except death and taxes" \cite{sparks_jared_letter_1856}. While robots may not have to pay taxes (yet \cite{Arduengo2021}), they must battle the same force that will lead to our death and the inevitable heat death of the universe: entropy \cite{Frautschi1982}. By robots simply acting in the world around them, entropy, be it friction in their joints or an errant particle of space radiation hitting their memory, makes system failure inevitable. 
Robots are increasingly integrated into critical applications, ranging from industrial automation to healthcare and even space exploration \cite{CensusSurveyRobotic2022, Kyrarini2021, papadopoulos2021robotic}. The reliability and continuous operation of these robotic systems are crucial for their acceptance and utilization in such demanding environments. However, as robots perform tasks over extended periods, the likelihood of mechanical failures increases \cite{austin2020robotic, She2016, muirhead2012coalescing}. A particularly critical and crippling case is when multiple joints fail simultaneously. These multi-joint failures can drastically reduce the robot's task space and challenge standard motion planning algorithms due to the constraints of the robot's configuration and control spaces \cite{kingston2018sampling}.

\begin{figure}
  \centering
    \includegraphics[width=\columnwidth]{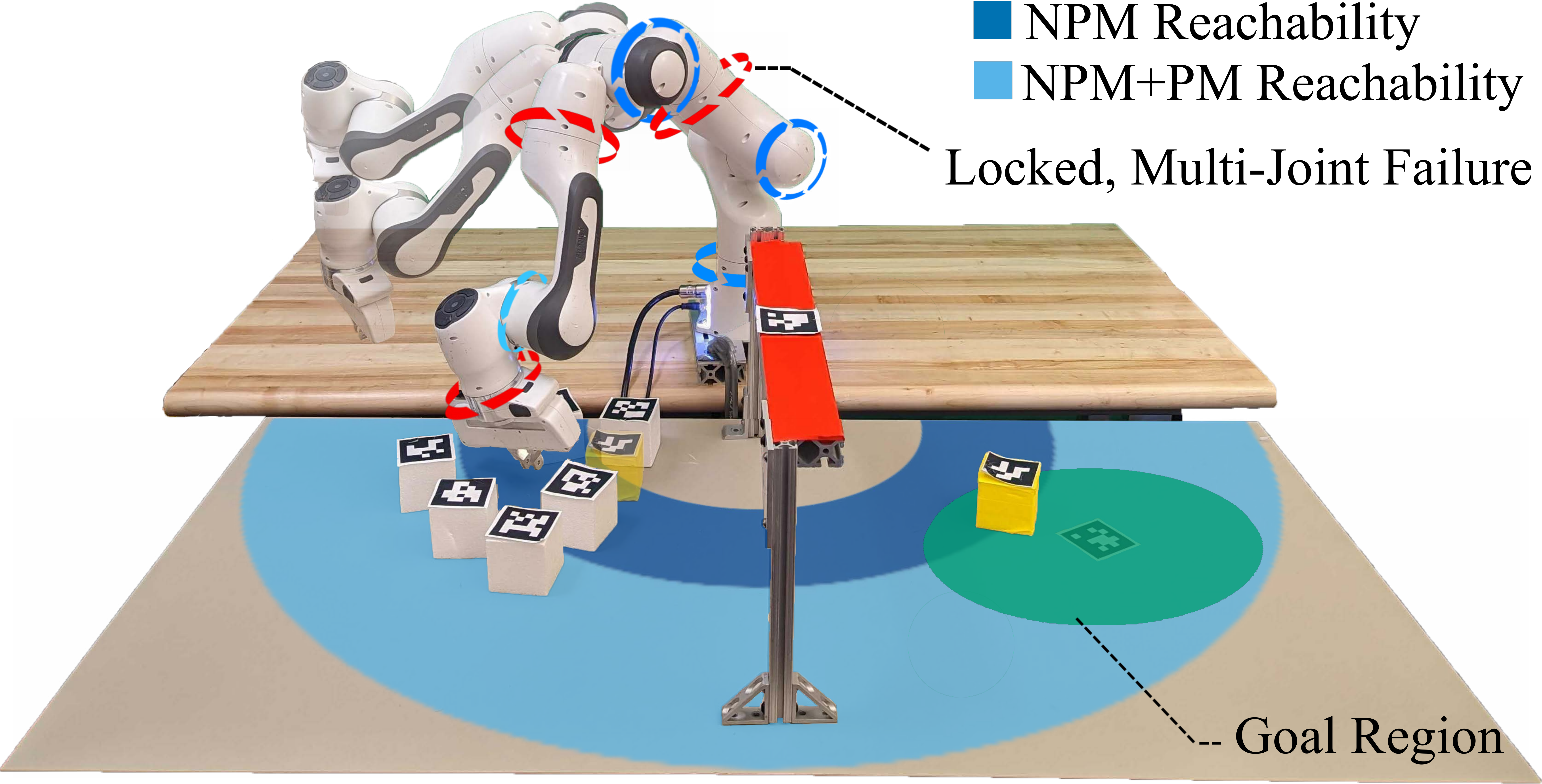}
  \caption{This figure illustrates a robot leveraging non-prehensile manipulation (NPM) to maintain manipulation capability despite experiencing locked, multi-joint (LMJ) failures (indicated by the red rings around the joints). The robot manipulates the yellow target object from its initial position to the goal region on the right within the failure-constrained workspace represented by the blue and green regions. }\label{fig:first-page}
\end{figure}

These constraints in the robot's kinematics and dynamics can degrade its manipulation capability through reduced abilities. In this work, we define ability as a motor skill a robot can recall to manipulate an object and capability as the extent to which a robot can complete a manipulation task. For a robot manipulator that becomes underactuated due to multi-joint failures, several constraints can occur: i) the position and orientation of the end effector become coupled to the robot's remaining functional joints, ii) the manipulable area will be constrained to a subset of its non-failure size, iii) grasping may not be possible due to task constraints, such as an object being in an ungraspable orientation, iv) the robot may lose the ability to manipulate the end-effector entirely due to an adverse configuration, and v) the end effector will be subject to nonholonomic constraints \cite{Bloch_Krishnaprasad_Murray_2016}. Thus, finding a motion to connect two task-space points will require significant time to calculate or learn \cite{atreya_steering}.

To counteract the limits multi-joint failures impose on the robot's manipulation capability, we must shift the manipulation paradigm to include abilities beyond grasping and interaction beyond the end-effector. We propose using non-prehensile manipulation (NPM), or abilities corresponding to ``anything but" grasping, through contact across the whole robot embodiment, to compensate for these constraints. As seen in \cref{fig:first-page}, these new abilities will increase the robot's reachable area and enable the completion of manipulation tasks after multi-joint failure.

We adapt kinodynamic edge bundles to store NPM actions in a failure-centric map \cite{shome2021asymptotically}, a method that captures the effects of nonholonomic constraints on end effector motion. This pre-computation approach enables multi-query planning, enhancing task planning efficiency. To utilize the kinodynamic map, we developed a sim-in-the-loop task planner. By simulating actions, the planner can estimate the interactions between the robot and the environment, improving the quality of the manipulation.

This research introduces a novel method to improve manipulation capability in the event of locked, multi-joint (LMJ) failures \cite{lewis1997fault, xie2021maximizing} by using NPM abilities. The contributions of this approach are: i) modeling the failure-constrained workspace and manipulation capability of the robot, ii) generating a kinodynamic map of NPM actions across the failure-constrained workspace, and iii) a manipulation planner that uses the kinodynamic map with physical interaction simulation. We introduce a new approach to enhancing the manipulation capabilities of robots, even in the face of locked, multi-joint failures.

The remainder of this paper is structured as follows: \cref{sec:related} discusses related work, and \cref{sec:methods} describes the approach for modeling LMJs and planning under these conditions using whole-arm and NPM actions. \cref{sec:eval-design} lays out how we evaluate our approach. \cref{sec:eval} presents the experimental evaluation results, where we assess the improvements in the manipulation area and demonstrate the capability to conduct tabletop manipulation tasks under two different LMJ failure cases. \cref{sec:conclusion} concludes the paper by discussing these findings and future work.

\section{Related Work}\label{sec:related}
Developing fault-tolerant systems is critical for autonomous robots that operate in harsh environments for extended periods \cite{muirhead2012coalescing, papadopoulos2021robotic}. This involves the ability to diagnose, recover from, and communicate failure modes \cite{visinsky1991fault, lewis1997fault, honig2018understanding}. While diagnosis and communication have been studied extensively, modeling and planning for autonomous failure recovery is underexplored. This is due to a limited, prehensile view of robotic manipulation \cite{Billard2019, SUOMALAINEN2022104224}. 
The proliferation of prehensile methods for robotic manipulation is primarily due to certainty in object pose once it is grasped. Prehensile-only approaches limit the scope of robot capability \cite{ruggiero2018nonprehensile}, necessitating the need for exploring NPM abilities.

\subsection{Failure Recovery}
To recover from failure, we must understand what effects a failure causes. Partial or total degradation resulting from extended operation and environmental factors may lead to a mismatch between task requirements and available system capability \cite{austin2020robotic, muirhead2012coalescing, shafaei2022dust}. 
The current literature has not thoroughly explored approaches to recovering from kinematic impairments in redundant robot manipulators.  The redundancies of robot manipulators can ensure recoverability for single-joint failures. However, multi-joint failures can severely reduce the robot's workspace to the point of task failure \cite{ porges2021planning}. Task success with joint failures is low using prehensile-only approaches but can be improved by considering non-prehensile modes of manipulation \cite{ruggiero2018nonprehensile}.

% The impact the failure has on the robot falls on a continuous spectrum of severity, from \textsl{completely irrecoverable} to \textsl{completely recoverable} \cite{vonasek2015online, xie2021maximizing, jamshidi2019machine}.

% Prior work mainly follows a limited scope of prehensile manipulation \cite{SUOMALAINEN2022104224}\gm{investigate citation}.
% grasping and end-effector.

Prior approaches to recovering from mechanical impairments include modeling the failures via analytical methods \cite{du2017fault, shome2021asymptotically} and using analytical inverse kinematics to explore and plan in the reachable space \cite{mu2016kinematic}. This requires time-consuming computation of the analytical models. Other work has developed approaches of constricting the motion plan of the robot before failure to maximize post-fault reachable space, \cite{xie2021maximizing, lewis1997fault}. 
In contrast, our approach does not need pre-failure constraints or analytical computation. This allows us to plan for an arbitrary permutation of multi-joint failures efficiently.

\subsection{Non-Prehensile Manipulation Modeling and Planning}

Non-prehensile manipulation in robotics has traditionally focused on isolated tasks or scenarios with redundant or fully-actuated robots \cite{ruggiero2018nonprehensile,mason2018toward}.
Past research has explored various NPM primitives, such as throwing \cite{zeng2020tossingbot}, pushing \cite{stuber2020let}, poking \cite{pasricha2022pokerrt}, catching \cite{kim2014catching}, flipping, and rolling \cite{lynch1999dynamic, ruggiero2018nonprehensile}, each with its own set of advantages and challenges. Yet, the integration of these primitives in under-actuated, LMJ-constrained robots remains unexplored.
%There exist a multitude of skills that fall in this domain, such as poking \cite{pasricha2022pokerrt}, pushing \cite{RSS2020Changkyu}, throwing \cite{zeng2020tossingbot}, flipping, and rolling \cite{lynch1999dynamic, ruggiero2018nonprehensile}. 

The likelihood of generating a valid plan over this severely constrained reachable space in an online manner, especially for dynamic actions such as poking, is near-zero \cite{kingston2018sampling}. This informs the need for precomputation of valid actions to guide planning for NPM primitives in the failure-constrained kinematic and control spaces \cite{westbrook2020anytime,chamzas2019using,shome2021asymptotically}.
% Leveraging this pre-generated map of actions to guide search becomes a key component for planning in high-dimensional configuration spaces, common in non-prehensile planning .
% Sampling from the constrained robot configuration space can be expensive, 
% so adaptive sampling on the relevant action subspaces may be combined with nearest neighbor search to reduce task times \cite{jentzsch2015mopl}.
To that end, we extend the concept of uniformly-sampled kinodynamic edge bundles \cite{shome2021asymptotically} to the failure domain by sampling edges using failure-constrained control inputs, boosting the likelihood of task success \cite{li2015sparse, kingston2018sampling}.
% on the failure manifold based on forward propagation of the robot dynamics.

% Previous work assumes uniform coverage of the configuration space in these edge bundles.
% However, this assumption will not hold for the failure constrained action space as the liklihood of sampling a feasible action approaches zero.

% Therefore, we introduce the generation of edge bundles by exploring parts of the control and configuration space that lie within the robot's failure-constrained reachable space.

% The prehensile view of robotic manipulation limits these approaches--grasping as a reliable skill for interacting with the environment and the end-effector as the sole point of contact between the robot and the surrounding environment \cite{king2015nonprehensile}.

% \input{adaptr_related}
\section{Methods}\label{sec:methods}
\begin{figure*}
    \centering
    \includegraphics[width=\textwidth]{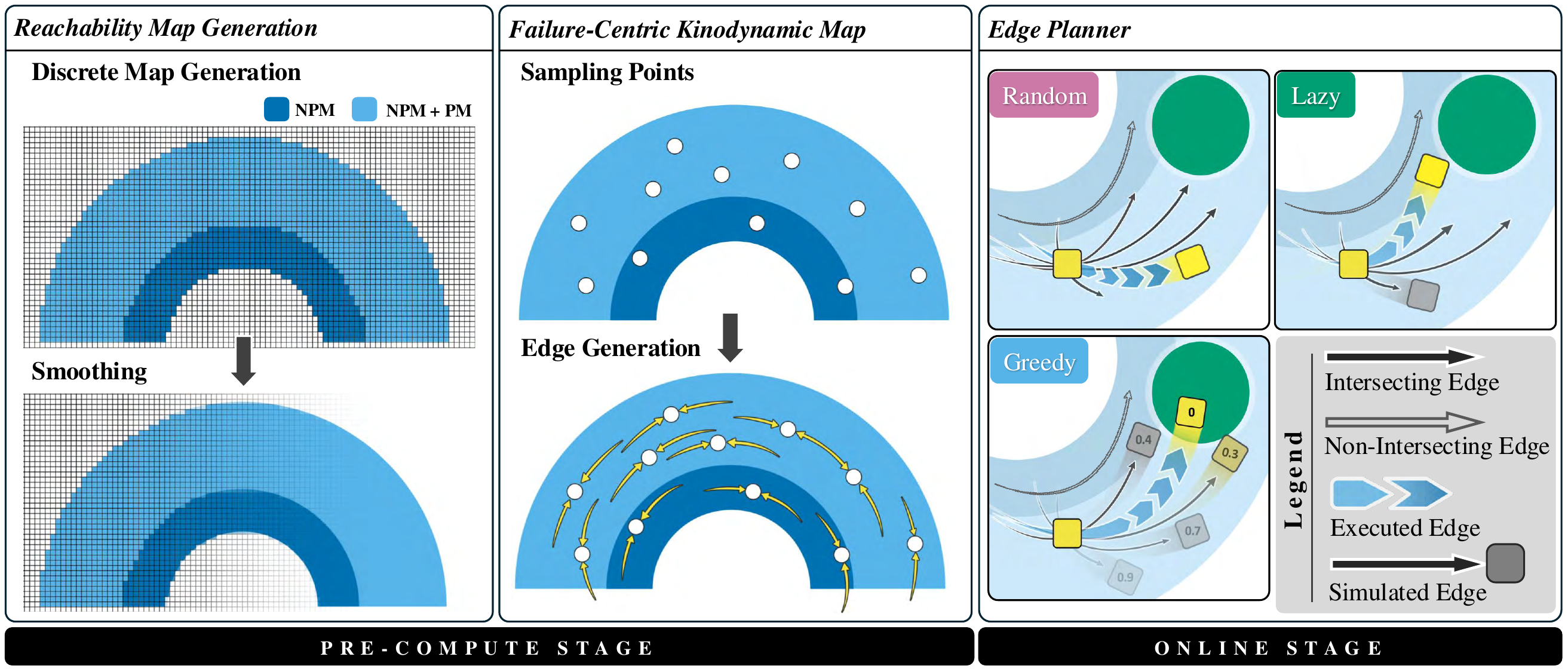}
    \caption{An overview of the approach to expand manipulation capability in a robot experiencing LMJs. We generate a reachability map that captures the manipulation capability via \cref{algo:reachability} and smooth it with interpolation. We use this reachability map to determine the failure-centric kinodynamic map, via \cref{algo:generate-kinodynamic-manifold}, by sampling points in the reachable space and then saving dynamics rollouts (yellow arrows) as edges. To use this kinodynamic map, we developed an edge planner that can vary its use of sim-in-the-loop to determine which action the robot will take via the Random, Lazy, and Greedy Action Selection Mechanisms in \cref{algo:planner}. }
    \label{fig:methods}
\end{figure*}

To enable robotic arms to maintain operational utility in the face of locked multi-joint failures, our method diverges from traditional paradigms by employing non-prehensile manipulation, particularly poking actions, to extend the operational capabilities of robots after failure (\cref{sec:methods-npm}). By redefining motion primitives within the robot's new kinematic constraints, we explore an expanded operational space through a strategy that includes the pre-computation of reachable states (\cref{sec:methods-reachability}, \cref{fig:methods}, Left Column) and the formulation of kinodynamic edge bundles for feasible action planning (\cref{sec:methods-modeling}, \cref{fig:methods}, Center Column). We developed multiple approaches to selecting actions to understand the benefits of including a simulation in the planning and execution loop (\cref{sec:methods-planning}, \cref{fig:methods}, Right Column).

\subsection{Expanding Robot Capabilities through Non-Prehensile Manipulation}\label{sec:methods-npm}

%The prehensile perspective on robotic manipulation constrains existing approaches in enabling continuous robot operation in the presence of LMJs

Existing manipulation approaches cannot complete tasks in the presence of LMJs as they are restricted to exclusively prehensile actions. They regard grasping as a dependable method for engaging with the surroundings and consider the end-effector as the exclusive point of interaction between the robot and its environment.
Our approach challenges these notions by utilizing whole-body NPM to enhance the probability of task completion. 

We leverage the poking primitive in our work as a representative NPM ability to enable manipulation in the presence of LMJs since it significantly expands the set of reachable configurations and the number of degrees of freedom of the robot's operational space. Prior work has defined poking as a fundamental nonprehensile motion primitive
that is composed of two sequential phases: i) impact: a robot
end-effector applies an instantaneous force to an object at
rest, setting it in planar translational and rotational motion;
and, ii) free-sliding: the object slows down and comes to
a halt due to Coulomb friction between the object and its
support surface.  This primitive is not limited by robot
reach and leads to faster manipulation of objects in dense clutter, in the presence of occlusion, and in ungraspable configurations. \cite{pasricha2022pokerrt}

The challenge of identifying a viable sequence of NPM actions, such as poking, for a robot is exemplified by the high computational demands seen in algorithms like \textit{PokeRRT} \cite{pasricha2022pokerrt}. This process requires the simulation of numerous action sequences, many of which may prove infeasible for a specific robot failure mode. Therefore, this uniformly random exploration of the manifold of valid states and control parameters is inherently inefficient.
The complexity is further compounded when dealing with robots experiencing locked joint failures. In such scenarios, the search must look beyond only the space of control parameters for our chosen motion primitives and require these actions to be feasible within the altered kinematic landscape imposed by the failure. This intersection often leads to a zero-measure manifold, which is difficult to sample when following a uniformly random strategy.
To mitigate this issue, we propose a pre-computation approach, which maps the reach and capability post-LMJ and then finds a set of dynamic actions the robot can achieve, considering the limitations imposed by LMJs. This precomputed map of kinodynamically feasible actions serves as a guide, significantly reducing the computational overhead by providing a reference for what is achievable given the current failure mode.

\begin{algorithm}\footnotesize
    \caption{Generate Reach-Ability Map}\label{algo:reachability}

    \KwIn{$\mathcal{W}$, Workspace of the Robot; $\mathcal{X}$, set of joint constraints; $k$ replan attempts; $\epsilon$, perturbation radius}
    \KwOut{$\mathcal{W}^c$, the Failure-Constrained Workspace of the Robot}
    $iter = 0$ \\
    \For{$\mathbf{p} \in \mathcal{W}$}{
        \textsc{InteractionType} = ``PM" \\
        \textsc{InteractionArea} = ``End Effector" \\
        success $\longleftarrow$ inverseKinematics($\mathbf{p}$, $\mathcal{X}$, \textsc{InteractionType}, \textsc{InteractionArea}) \\
        \While{iter $< k$}{
            \If{iter $> 1$}{
                $\mathbf{\rho} \longleftarrow$ \textsc{SampleUniformEpsilonBall}($\epsilon$) \\
                $\mathbf{p}_{\rho} \longleftarrow \mathbf{p} + \mathbf{\rho}$ \\
                \If{iter $>k/2$}{
                \textsc{InteractionType}, \textsc{InteractionArea} $\longleftarrow$
                \textsc{ChangeInteraction}() }
                success $\longleftarrow$ \textsc{InverseKinematics}($\mathbf{p}_{\rho}$, $\mathcal{X}$, \textsc{InteractionType}, \textsc{InteractionArea})
            }            
            \If{success}{
                $\mathcal{W}^c \longleftarrow \mathcal{W}^c \cup \mathbf{p}$
                \break
            }
            $iter \longleftarrow iter + 1$ 
        }
    }
    \Return{$\mathcal{W}^c$}
\end{algorithm}

\subsection{Finding Reach and Manipulation Capability After Failure}\label{sec:methods-reachability}
When an LMJ occurs, the bounds of the robot's reachable workspace change in intractable, non-linear ways. We must understand the new reachable area and discover what manipulation capabilities the robot retains across this failure-constrained region to continue to complete tasks.

To do this, we developed a reachability and ability analysis system, or \textit{Reach-Ability} (\cref{algo:reachability}). The task space of the robot is divided into small areas approximately the size of the Franka Emika Panda end-effector tip, $2$ cm (\cref{fig:methods}, Left Column, Top). For each discretized area, we check to see if the robot can find an inverse kinematics solution for prehensile actions. To be considered a valid prehensile pose, the end-effector must be in a range of orientations that allow for grasping, nominally between normal to the workspace surface and parallel to the workspace surface. To counter the effects of the randomness of the inverse kinematics solver, if no solution is found initially, we uniformly sample a perturbation from an $\epsilon$-ball inscribed within the discretization cell and attempt solving again. 

If there is still no solution after half the replan attempts are calculated, \texttt{changeInteraction()} (\cref{algo:reachability}, L9) will release the orientation constraints of PM manipulation and switch to NPM abilities.%change the interaction type from PM to NPM
Under the NPM abilities, the inverse kinematics solver (\cref{algo:reachability}, L10) will: 1) cycle through pre-selected interaction points on the robot: hand, wrist, and forearm and 2) relax its orientation error bound so that the solution is only restricted in position. If no solutions are found, then that point is considered unreachable. The map is smoothed through interpolation once the whole task space is explored (\cref{fig:methods}, Left Column, Bottom).  
We leverage this mapping of the robot's statically reachable, failure-constrained workspace to build a representation of the robot's action manifold by utilizing the notion of edge bundles that parameterize the set of all feasible actions at the intersection of the post-failure-accessible manifold and the motion primitive manifold. 
% This still doesn’t allow for the execution of actions which leads us to a hierarchical representation - a notion of edge bundles that parameterize the set of all feasible actions at the intersection of the failure manifold and the motion primitive manifold. Brute force iteration.

% an understanding of the robot's failure-constrained workspace, we can now construct the kinodynamic map within those bounds.

\subsection{Construction of the Failure-Centric Kinodynamic Map}\label{sec:methods-modeling}

% \begin{itemize}
%     \item Introduce the notion of robot joint space manifold, $\mathcal{C}_q$
%     \item corresponding control space manifold $\mathcal{C}_u$
%     \item when a failure occurs, these manifolds are constrained, becoming $\mathcal{C}_q_c $ and$ \mathcal{C}_u_c$, respectively
%     \item $\mathcal{C}_q_c$ is constrained so much that finding valid plans via uniform random sampling and connecting them in $\mathcal{C}_u_c$ is unfeasible
%     \item so we use edge bundles to approximate $\mathcal{C}_q_c$ by forward propagating the robot's dynamics equation for a valid sampled control input in $\mathcal{C}_u_c$
%     \item each of these rollouts $r_q$ is saved in $\mathcal{E}$
% \end{itemize}

We can represent the set of all possible joint states and valid control inputs as the manifolds $\mathcal{C}$ and $\mathcal{U}$, respectively. We note the end-effector workspace of the robot as $\mathcal{W}$. When an LMJ occurs, these manifolds become degenerate, failure-constrained spaces. We note these constrained manifolds as $\mathcal{C}_c$, $\mathcal{U}_c$, and $\mathcal{W}_c$, respectively. 

The mapping between these spaces, $\mathcal{C} \mapsto \mathcal{C}_c$ and $\mathcal{U} \mapsto \mathcal{U}_c$, is nonlinear and dependent on the failure configuration of the robot. They also lose $n$ dimensions for $n$ locked joints. This degeneration means the failure-constrained manifolds have a significantly smaller volume than the unconstrained manifolds. As a result, the probability of finding a valid connecting plan between two uniformly, randomly sampled points in $\mathcal{C}_c$, with control inputs sampled from $\mathcal{U}_c$, approaches zero. \cite{kingston2018sampling}.

% Assuming a valid diagnosis of failure modes and the corresponding joint values for the locked joints, we follow a joint angle discretization approach on the remaining movable joints to map the robot's kinematically reachable space. 
Theoretically, an analytical solution to the inverse kinematics exists for any given point in $\mathcal{W}_c$. However, the multiple possible permutations of failures across the joints mean deriving the relevant equations or sampling valid configurations is untenable. As a result, we approximate $\mathcal{C}_c$ through a discretization-based approach.

We utilize the notion of edge bundles from \cite{shome2021asymptotically}. These edge bundles, $\mathcal{E}$, are a collection of Monte Carlo rollouts, or edges, $\mathbf{e}$ of the robot dynamical model with random control inputs, $\mathbf{\Dot{q}} \in \mathcal{U}_c$.
The nature of these edges, i.e., simulating a random control input from a random state for a random duration, ensures probabilistic completeness of the planning algorithm in \cref{sec:methods-planning} \cite{kleinbort2018probabilistic}. 
% (Note that claims on completeness may be inconclusive with respect to our planner since it uses the best-first input approach \cite{kleinbort2018probabilistic}.)
Following the procedure outlined in \cref{algo:generate-kinodynamic-manifold}, we use this Monte Carlo generation of edge bundles to produce a \textsl{failure-centric kinodynamic map} (\cref{fig:methods}, Center Column).
In practice, the set of all edges, $\mathcal{E}$, provides a discrete approximation to $\mathcal{C}_c$.

\begin{algorithm}\footnotesize

\caption{Generate Kinodynamic Manifold.}\label{algo:generate-kinodynamic-manifold}

\KwIn{$\mathcal{C}^c$, Constrained Configuration Space; $\mathcal{U}^c$, Constrained Control Space; $n$, number of samples; $\mathcal{W}^c$, failure-constrained workspace; $\Delta t$, the control timestep of the robot}
% ; $m_{min}$, the minimum acceptable manipulability measure
\KwOut{$\mathcal{E}$, a set of edges}

$\mathcal{E} \longleftarrow \emptyset$

\For{$i = 1 \ldots n$} {
    $\mathbf{e} \longleftarrow \emptyset$ \tcp{edge}
    $\xi_{EE} \longleftarrow$ \textsc{SamplePoint}($\mathcal{W}^c$) \tcp{XYZ Only}
    $\mathbf{q} \longleftarrow$ \textsc{PositionalInverseKinematics}($\xi_{EE}, \mathcal{C}^c$) \tcp{Orientation Unconstrained}
    $t \longleftarrow$ \textsc{SampleDuration}()\\
    $\mathbf{\dot{q}} \longleftarrow$ \textsc{SampleControlInput}($\mathcal{U}^c$) \\
    $valid \longleftarrow True$ \\
    \tcp{simulate}
    \While{\textsc{TimeRemaining}($t$)} {
        % $\dot{\mathbf{\dot{q}}} = \mathbf{0}$ \\ \tcp{const vel}
        $\ddot{\mathbf{q}} \longleftarrow J(\mathbf{q})^\dagger \times (\dot{\mathbf{\dot{q}}} - \dot{J}(\mathbf{q}))\mathbf{\dot{q}})$\\
        $\mathbf{\tau} \longleftarrow M(\mathbf{q})\ddot{\mathbf{q}} + C(\mathbf{q}, \mathbf{\dot{q}})\mathbf{\dot{q}} + g(\mathbf{q}) + f(\mathbf{\dot{q}})$\\
        $m \longleftarrow \sqrt{\det{J(\mathbf{q})J(\mathbf{q})^T}}$ \\
        \If{\textsc{LimitsViolated($\mathbf{q}$, $\ddot{\mathbf{q}}$, $\mathbf{\tau}$)} \textbf{or} \textsc{InCollision}($\mathbf{q}$)} 
        % \textbf{or} $m < m_{min}$
        { 
            $valid \longleftarrow False$\\
            \textbf{break}
        }
        $\mathbf{q}_{k+1} \longleftarrow \mathbf{q}_{k} + \mathbf{\dot{q}} \Delta t$ \\
        $\mathbf{e} \longleftarrow \mathbf{e} \cup \{\mathbf{q}_k, \mathbf{\dot{q}}\}$
    }
    \If{valid} {
            $\mathcal{E} \longleftarrow \mathcal{E} \cup \mathbf{e}$
    }

}
\Return{$\mathcal{E}$}

\end{algorithm}

A kinodynamic map is computed once offline for a given failure. Since NPM interactions are inherently stochastic, we explicitly only model the robot's dynamics in our edge bundle. These attributes allow reuse across many planning problems, making our motion planner multi-query.

% \subsection{Planning for Failure Recovery}\label{sec:methods-planning}
\subsection{Motion Planning under LMJ Failure Conditions}\label{sec:methods-planning}

The \textsl{failure-constrained kinodynamic map} provides a set of actions for our planner to complete the manipulation task. It provides two main advantages in the context of failure-constrained motion planning: i) it removes the need for an inverse model such that no action sampling is required during the planning process, and ii) it caters to a multi-query planning framework to speed up planning times for the failure-case significantly. 
% \gm{add reason why multi-query is good}

\begin{algorithm}\footnotesize
\caption{Edge Planner.}\label{algo:planner}
\KwIn{$E$, The Environment State; $\mathcal{E}$, a set of edges; $m$, Action Selection Method Mode}
\KwOut{$\mathcal{A}$, an action for the robot to execute}

$\mathcal{T} \longleftarrow \emptyset$
% \textsc{TimeRemaining}() \textbf{or} 

\While{\textbf{not} \textsc{GoalReached($E$)}} {
    % node = \textsc{PickNode}()\\
    % \textsc{ResetEnvToState}(node)\\
    $\mathcal{E}_I \longleftarrow$
    \textsc{EdgesIntersectingObject}($\mathcal{E}$, $E$)\\
    \color{dusty_pink}
    \If{m=``Random"}{
        $e \longleftarrow \textsc{RandomUniformSelect}(\mathcal{E}_I)$\\
        $\mathcal{A} \longleftarrow e$ \\
        \Return{$\mathcal{A}$}
    }
    \color{dusty_green}
    \If{m=``Lazy"}{
        $E_{sim} \longleftarrow E $ \tcp{set sim env}
        \For{$\mathbf{e}$ in \textsc{SampleWithoutReplacement}($\mathcal{E}_I$)} {
            \textsc{SimulateRobotWithEdge($\mathbf{e}, E_{sim}$)} \\
            \If{\textsc{TargetMovedTowardsGoal}($E_{sim}$)}{
            $\mathcal{A} \longleftarrow \mathbf{e}$\\
            \textbf{break}
            }
        $E_{sim} \longleftarrow E$ \tcp{reset sim env}
        }
        \Return{$\mathcal{A}$}
    }
    \color{dusty_blue}
    \If{m=``Greedy"}{
        $\mathcal{A} \longleftarrow \emptyset$\\
    
        $\mathcal{E}_T \longleftarrow \emptyset$ \tcp{Edges Scored}
        $scores \longleftarrow \emptyset$ \\
        $\mathcal{E}_s \longleftarrow \textsc{UniformlySampleSubset}(\mathcal{E}_I)$ \\
        $E_{sim} \longleftarrow E $ \tcp{set sim env}
        \For{$\mathbf{e}$ in \textsc{SampleWithoutReplacement}($\mathcal{E}_s$)} {
            \textsc{SimulateRobotWithEdge($\mathbf{e}, E_{sim}$)} \\
            \If{\textsc{TargetMovedTowardsGoal}($E_{sim}$)}{
            $scores \longleftarrow scores \cup \textsc{ScoreExecution}(E_{sim}$)}
        $\mathcal{E}_T \longleftarrow \mathcal{E}_T \cup \mathbf{e} $

        $E_{sim} \longleftarrow E$ \tcp{reset sim env}
        }
         
        $\mathcal{A}\longleftarrow \textsc{SelectBestEdge}(\mathcal{E}_T, scores, m)$\\
        % $ \longleftarrow e_{best}$ \\
        % }
        \Return{$\mathcal{A}$}
    }
    % }
}

\end{algorithm}

Based on the approximations of the configuration and control manifolds ($\mathcal{C}_c$ and $\mathcal{U}_c$), our use of nonprehensile actions and environmental contact with the full embodiment of the robot arm enhances task success in the presence of LMJs.
Given a start configuration of the scene $x_0 \in E$, the motion planning for the failure recovery problem is to find a trajectory $\tau: [0, 1] \rightarrow E_{free} \cap \mathcal{W}_c$ such that $\tau(0) = x_0$ and $\tau(1) \in E_{goal}$ (\cref{fig:methods}, Right Column). Success is manipulating the target object from its starting pose to the goal region (\cref{algo:planner}).
% We specifically leverage the concept of multiple modalities across PM and NPM skills to accomplish manipulation tasks in the face of LMJ. For example, if the end-effector joints are locked and positioned in a way such that the robot cannot grasp, as shown in \cref{fig:panda_crippled}, we take advantage of poking to interact with the object. 

While the failure-constrained kinodynamic manifold captures the robot dynamics, interactions with scene objects are unmodeled because physical properties, especially friction for NPM interactions, behave stochastically in the real--world \cite{ALBERTINI2021104242}. However, physics engines provide approximately realistic predictions of rigid body dynamics. We use the Bullet physics simulation engine in the planning loop to get the resultant pose $x_{k+1}$ \cite{coumans2021}.  This allows us to capture an approximation of robot and object dynamics faster and safer than execution on a real robot. 
% Each action execution follows a validity check to ensure all objects remain within workspace bounds and no collisions occur between movable objects and fixed obstacles (\Cref{algo:multimodal-planner}, L8).

% Our planner interleaves planning and execution to compensate for discrepancies between simulation and real-world dynamics (\cref{algo:adaptr}).
To understand the benefits of utilizing a simulator in the planning loop, we study multiple Action Selection Mechanisms (ASMs):
\begin{itemize}
    \item \textbf{\color{dusty_pink}Random} (\cref{algo:planner}, L4-7): The Random ASM finds all the edges that intersect the object and picks one randomly.
    \item \textbf{\color{dusty_green}Lazy} (\cref{algo:planner}, L8-16): The Lazy ASM utilizes our sim-in-the-loop planner to find an edge that moves the target object toward the goal. Of all the edges that intersect the object, it picks the first action that moves the target towards the goal in the simulator.
    \item \textbf{\color{dusty_blue}Greedy} (\cref{algo:planner}, L17-30): The Greedy ASM will sample a subset of the edges that intersect the object and score them based on how close the target object is to the goal after simulating the selected actions. It will command the highest-scored edge for the real robot.
\end{itemize}

\section{Experimental Design}\label{sec:eval-design}
% \begin{figure}
%     \centering
%     \includegraphics[width=\columnwidth]{figures/planner_ops_smol.png}
%     \caption{The Planning Procedure: Step 1: Using the method from \cref{fig:edge_selection}, select edges to try in simulation; Step 2: Try the edges in simulation and save the action that moves the object closest to the goal; Step 3: Execute that edge on the real robot.}\vspace{-15pt}
%     \label{fig:timeline}
% \end{figure}
\begin{figure}
    \centering
    \includegraphics[width=\columnwidth]{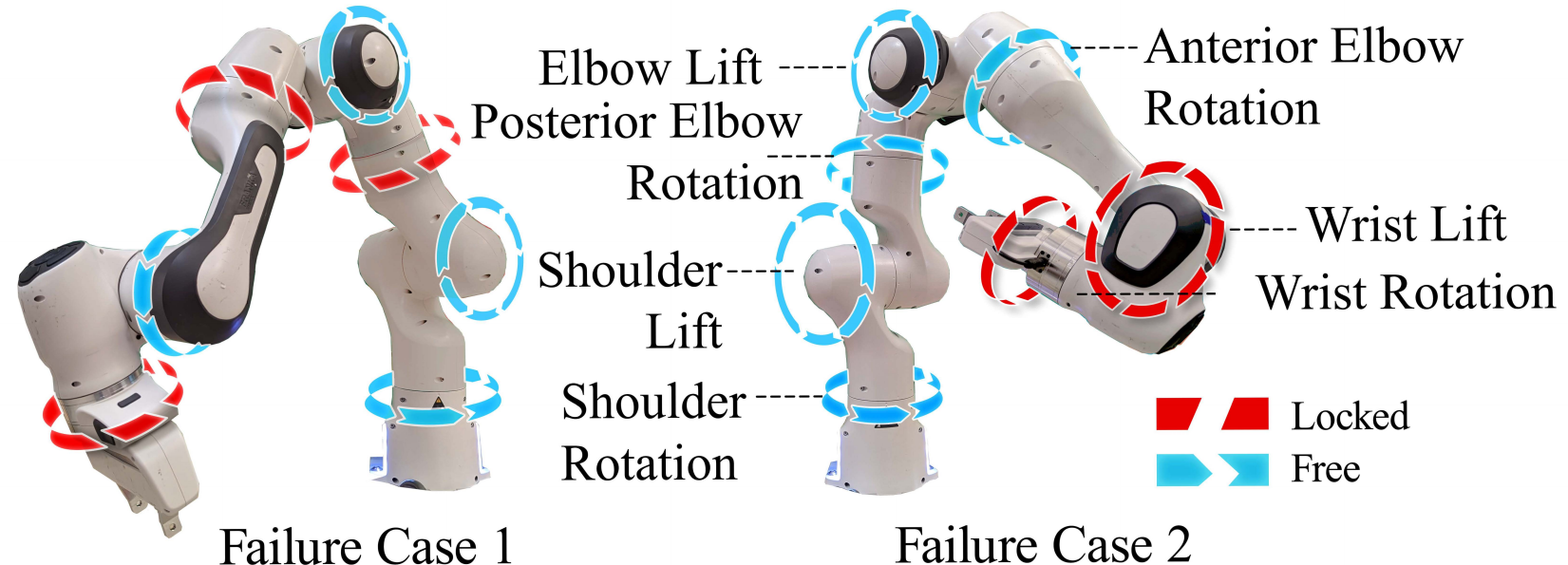}
    
    \caption{Franka Emika Panda robotic arm with multiple locked joints. The left represents Failure Case 1, with the locked Anterior and Posterior Elbow Rotation and Wrist Rotation joints locked. The right represents Failure Case 2 with locked Wrist Lift and Rotation joints.}
    \label{fig:panda_crippled}
\end{figure}
\begin{figure}
    \centering
    \includegraphics[width=\columnwidth]{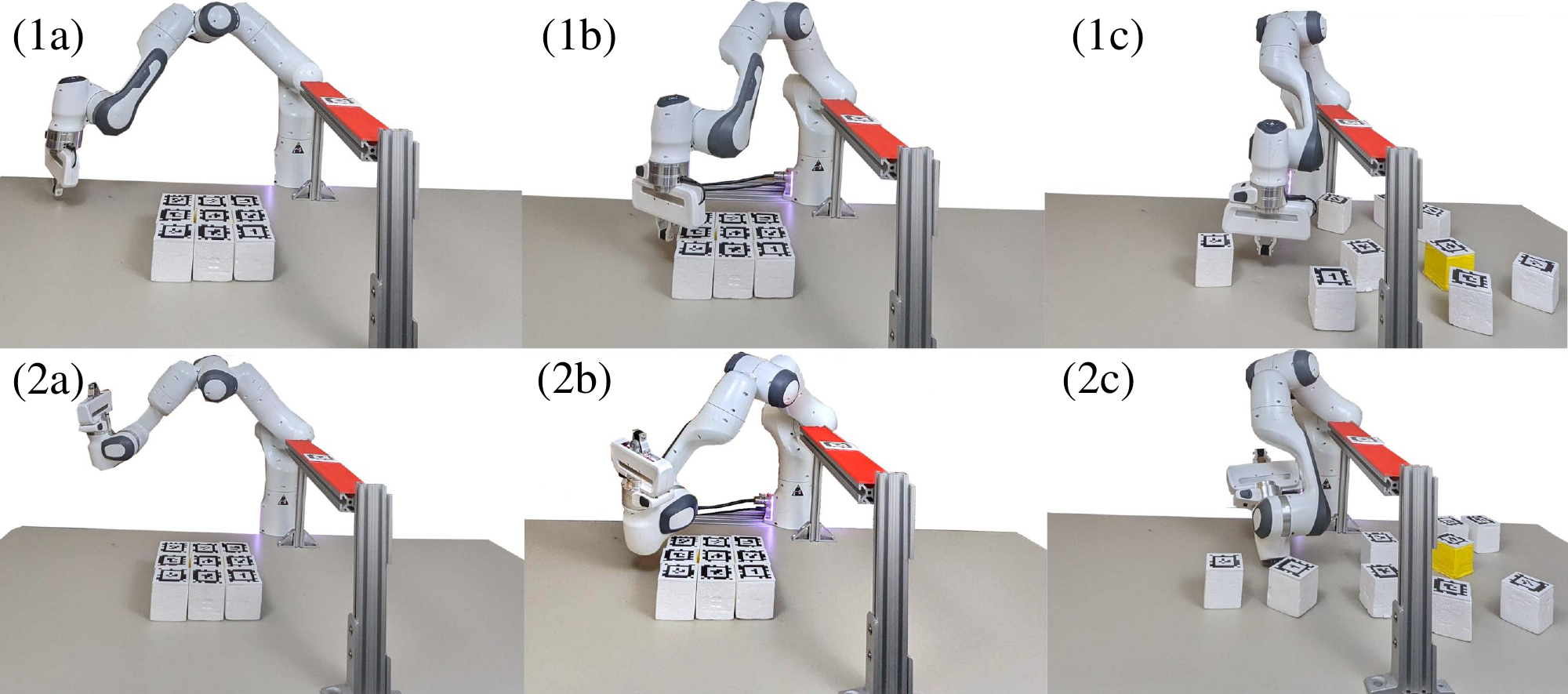}
    
    \caption{This figure presents two action sequences demonstrating the NPM interaction using different robot embodiment areas. The top sequence (1a-c) illustrates the interaction using the end effector, while the bottom sequence (2a-c) shows the interaction utilizing the robot’s whole embodiment. These sequences represent a composite scenario derived from scenarios 2 and 3, providing an illustrative view of the robot’s expanded capabilities.}
    \label{fig:action_example}
\end{figure}
\begin{figure*}
    \centering
    \includegraphics[width=\textwidth]{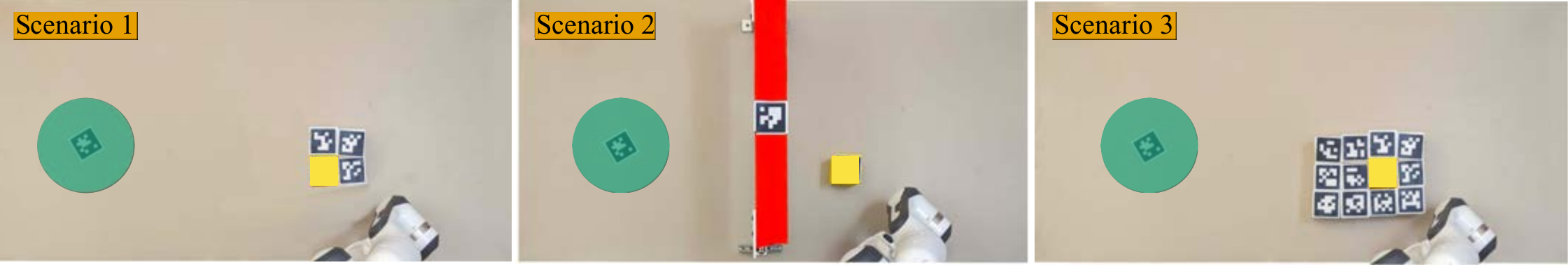}
    \caption{This figure presents three distinct scenarios tested in the study. Scenario 1 (left) features three movable obstacles (white with fiducial markers) and one target object (yellow). Scenario 2 (center) introduces a tunnel obstacle partitioning the table. Scenario 3 (right) includes 11 movable obstacles and one target object. The green shaded area represents the goal for each Scenario. \vspace{-15pt}}
    \label{fig:scenarios}
\end{figure*}
We test our system in three real-world scenarios with two LMJ cases across three different ablative ASMs. These failure cases, shown in \cref{fig:panda_crippled}, demonstrate two different families of failures that can occur: Case 1 limits the end effector grasping ability to a portion of the reachable area, and Case 2 precludes the use of the end effector in its entirety. Without loss of generality, these two cases represent a subset of the many permutations of faults that can occur due to LMJs. Our results are a quantitative analysis of the planning time, success rate, and reachability improvements resulting from utilizing the robot's whole embodiment combined with NPM actions. Two representative NPM actions are shown in \cref{fig:action_example}.
% \textit{ADAPT-R} enables the robot to complete tasks in these failure cases where prior approaches would fail \cite{xie2021maximizing, porges2021planning}.

We consider two failure cases in our evaluation: 
% The Failure Cases are as follows:
\begin{itemize}
    \item \textbf{Failure Case 1} (\cref{fig:panda_crippled}, left): The first case involves locking three joints: the Posterior Elbow Rotation, Anterior Elbow Rotation, and Wrist Rotation Joints.
    % This limits the robot's PM capability by drastically restricting the set of valid grasp configurations, as reflected by robot's reachable task space area and by limiting the end-effector orientations.
    This limits the graspable area and reduces the range of grasping configurations by coupling the end-effector orientation to the joint configuration.
    \item \textbf{Failure Case 2} (\cref{fig:panda_crippled}, right): In the second case, two joints are locked: the wrist lift and wrist rotation joints. This failure case precludes the use of the end-effector in completing manipulation tasks. This enables us to test the system using exclusively the non-end-effector parts of the robot's embodiment. 
\end{itemize}
The experimental scenarios are presented in \cref{fig:scenarios}. Each scenario involves manipulating the target object (the yellow cube) to the goal location (green area):
\begin{itemize}
    \item \textbf{Scenario 1} (\cref{fig:scenarios}, left): This scenario has three movable obstacles and one target object. This scenario is motivated by the need to be able to manipulate in cluttered environments that hinder direct access to the target object. 
     \item \textbf{Scenario 2} (\cref{fig:scenarios}, center): The second scenario has a tunnel obstacle in the middle, partitioning the two halves of the table. This scenario is motivated by the need to perform dexterous manipulation while maneuvering around environmental obstacles.
    \item \textbf{Scenario 3} (\cref{fig:scenarios}, right): There are 11 movable obstacles and one target object. This scenario allows us to explore the utility of our planner in the presence of increased object-obstacle interactions.
\end{itemize}

% We conduct an ablative study across multiple Pick Action Methods:
% \begin{itemize}
%     \item \textbf{Greedy}: The Greedy action selection method utilizes our sim-in-the-loop planner to find an edge that moves the target object toward the goal. Of all the edges that intersect the object, it selects the first action that moves the target towards the goal in the simulator.
%     \item \textbf{Random}: The Random action selection method finds all the edges that intersect the object and picks one randomly.
%     \item \textbf{Cautious}: The Cautious action selection will sample five percent of the edges that intersect the object and score them based on how close the target object is to the goal after simulating the selected actions. It will command the highest-scored edge for the real robot.
% \end{itemize}

% because our planner’s primary purpose is to complete tasks under failure.
To comprehensively understand how NPM abilities expand the robot's task space, we conduct the reachability analysis in \cref{sec:reach_anls}. In \cref{sec:planner_eval}, we show the effectiveness of our system across the three scenarios shown in \cref{fig:scenarios}, for the two failure cases described in \cref{fig:panda_crippled}, using the three Action Selection Mechanisms described in \cref{sec:methods-planning} over a total of 267 trials. This analysis shows the limitations of grasping-only actions and demonstrates the improvements brought about by NPM abilities.
% to show the limits in the task space of grasping-only actions and the improvements enabled through NPM abilities.
% Below, we present our findings on the increase in reachability for a given task plane in \cref{sec:reach_anls} and an analysis of our experimental results in testing \textit{ADAPT-R} in \cref{sec:planner_eval}. 
Through these results, we seek to verify two hypotheses:
% \begin{itemize}
%     \item[\textbf{H.1}] Combining NPM abilities with whole-body interaction increases the manipulable area of a robot arm experiencing LMJs.
%     \item[\textbf{H.2}] Implementing NPM abilities can increase the success of a robot conducting tabletop manipulation tasks.
% \end{itemize}
\begin{itemize}
    \item[\textbf{H.1}] When experiencing LMJs that limit or disallow the use of the end-effector, whole-body NPM abilities increase the manipulable area of a robot manipulator.
    \item[\textbf{H.2}] When experiencing LMJs that limit or disallow the use of the end-effector, whole-body NPM abilities can increase the success rate of a robot conducting tabletop manipulation tasks.
\end{itemize}

\section{Experimental Results}\label{sec:eval}

\subsection{Reachability and Manipulation Capability Analysis}\label{sec:reach_anls}
Relying solely on the end effector renders the robot ineffectual in failure cases that severely or entirely limit its use.
NPM abilities and the capability to interact with the environment using the whole embodiment allow for increased robot manipulable area. To study this, we analyzed the robot's manipulable area and capabilities over the task space of the evaluation scenarios.
The manipulable area is found through the method presented in \cref{sec:methods-reachability}.

\begin{table}[]
\centering
\begin{tabular}{@{}cccc@{}}
\toprule
\textit{\textbf{Failure Case}} & \textit{\textbf{Ability}} & \textit{\textbf{Area (m$^2$)}} & \textit{\textbf{Change in Reachable Area}}\\ \midrule
\textit{No Failure} & \textbf{NPM+PM} & \textbf{0.92} & Datum \\ %\midrule \textit{No Failure}
 & PM     & 0.85 & -$8\%$ \\ \midrule
\textit{1}       & \textbf{NPM+PM} & \textbf{0.85} & \textbf{-8\%} \\ %\midrule % 0.8521 \textit{1} 
      & PM     & 0.62 & -$33\%$ \\ \midrule
\textit{2}       & \textbf{NPM+PM} & \textbf{0.73} & \textbf{-21\%}  \\ %\midrule \textit{2}  
     & PM     & 0 & -$100\%$      \\ \bottomrule
\end{tabular}
\caption{Reachability Data \vspace{-15pt}}
\label{tab:reachability}
\end{table}

The nominal full-dexterity manipulable area is 0.85 $m^2$; the use of NPM increases this area to 108\% of nominal (\cref{tab:reachability}, rows 1 and 2). The manipulable area in the first failure case is severely limited by the inability to control the end-effector's orientation and task space movements in the z-axis when the end-effector is closer to the robot's base. When using NPM actions, the robot maintains 92\% of the reachable area compared to 67\% when using PM actions only (\cref{tab:reachability}, rows 3 \& 4). Since Failure Case 2 precludes the use of the end-effector, the PM-only reachable area is eliminated. However, using NPM actions, the reachable area is only reduced by 21\% (\cref{tab:reachability}, rows 5 \& 6).

The results show that \textbf{in failure modes that disallow or limit the use of the end effector, utilizing the whole body of the robot with NPM actions significantly expands the area of the robot's reachable space, supporting H.1.}

\begin{figure*}
    \centering
    \includegraphics[width=\textwidth]{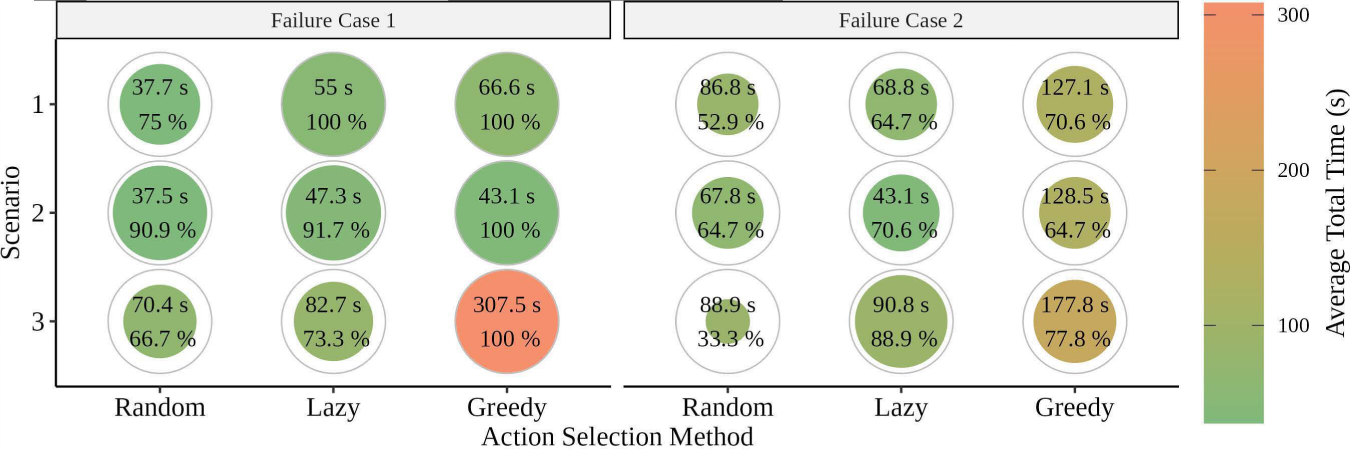}
    \caption{Comparative performance of Action Selection Methods across the Failure Cases. The color intensity of each circle indicates the task time, with green colors representing lower times and red colors indicating higher times. The size of each circle represents the success rate, with larger circles denoting higher rates.\vspace{-10pt}}
    \label{fig:bubble}
\end{figure*}
\begin{figure*}
    \centering
    \includegraphics[width=\textwidth]{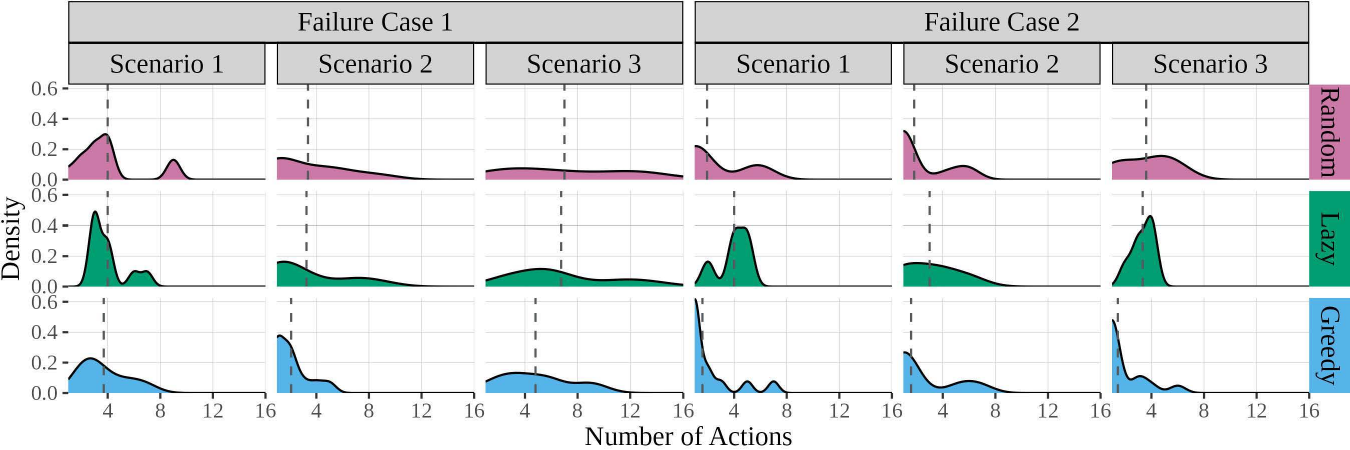}
    \caption{The distribution of actions required to complete different scenarios for each failure case using the action selection methods. Each graph represents a scenario under a specific failure case, with the x-axis indicating the number of actions and the y-axis representing the density across all trials. \vspace{-10pt}}
    \label{fig:shoulder-actions-comp}
\end{figure*}
\subsection{Real-world Manipulation Results}\label{sec:planner_eval}
% \begin{table}[]
% \centering
% \begin{tabular}{ccccc}
% \hline
% \textbf{}       & \multicolumn{3}{c}{\textbf{Task Time {[}seconds{]}}}                        & \textbf{Success Rate} \\ \hline
% \textbf{Planner}  & \textbf{S1} & \textbf{S2} & \textbf{S3} & \textbf{S1-S3} \\
% \textbf{Lazy} & \textit{\textbf{69.53}} & \textit{\textbf{62.25}} & \textit{\textbf{85.92}} &\textit{ \textbf{75.0\%}}       \\
% \textbf{greedy} & 146.23      & 130.46      & 127.53      & 61.2\%         \\
% \textbf{Random}   & 97.37       & 74.26       & 94.64       & 60.0\%         \\ \hline
% \end{tabular}
% \caption{Task Times For Failure Case 2}
% \label{tab:task-times-fc2}
% \end{table}

\cref{fig:bubble} shows the average total task time and success rates across Scenarios 1-3, Failure Cases 1 and 2, and the Random, Lazy, and Greedy ASMs. Across the ASMs, we see a trend of the Greedy method taking the longest, up to 307.5s, and the Random method taking the shortest time, as low as 37.5s, to complete the task, with the Lazy method roughly falling in between, at 47.3s-82.7s. This is expected because when averaged over all scenarios, the Greedy edge selection takes more time to plan than the other methods. Regarding the success rate, we see a clear trend in Failure Case 1, where the Greedy edge selection method has a 100\% success rate across all scenarios, and the Random edge selection has the lowest success rate. The Lazy edge selection has a success rate that, at best, matches the Greedy selection method at 100\% and, at worst, is better than the Random selection method, at 73.3\% vs. 66.7\% for Scenario 3. However, for Failure Case 2, no clear trend for the success rate occurs. Overall, the Lazy selection case is successful in scenarios 2 and 3, while the other two ASMs tend to have poorer results, except in scenario 1 for the Greedy ASM at 70.6\%. These results come from the combination of the inherent stochastic nature of the NPM interactions multiplied by the decreased sim-to-real accuracy of our sim-in-the-loop planner when interacting with areas of the robot that are not the end effector, as this failure case requires. This can partly be explained by the slight differences between our simulated collision meshes and the real robot embodiment.

% \begin{table}[]
% \centering
% \resizebox{\columnwidth}{!}{%
% \begin{tabular}{@{}ccccccc@{}}
% \toprule
%                   & \multicolumn{6}{c}{\textbf{Execution Time}} \\ \midrule
% \textbf{}                                              & \multicolumn{3}{c}{\textbf{Failure Case 1}} & \multicolumn{3}{c}{\textbf{Failure Case 2}} \\
% \multicolumn{1}{l}{\textbf{Selection Method}} & \textbf{Scenario 1}    & \textbf{Scenario 2}   & \textbf{Scenario 3}   & \textbf{Scenario 1}    & \textbf{Scenario 2}   & \textbf{Scenario 3}   \\
% \textbf{greedy} & 26.7  & 11.7  & 20.1  & 10.5  & 12.3 & 13.0 \\
% \textbf{Lazy}   & 27.1  & 19.7  & 21.2  & 13.7  & 10.1 & 21.1 \\
% \textbf{Random}   & 25.2  & 16.1  & 15.1  & 9.29  & 11.1 & 18.8 \\ \bottomrule
% \end{tabular}%
% }
% \caption{Execution Time Summary}
% \label{tab:exec-times}
% \end{table}

% \begin{figure*}
%     \centering
%     \includegraphics[width=\textwidth]{figures/time_comparison.pdf}
%     \caption{Comparison of Planning and Execution times across the Failure Cases and the Edge Selection Methods averaged over all Scenarios \gm{TODO: Fix Font and Text Size | Should this just be replaced with a table of averages?}}
%     \label{fig:time-comp}
% \end{figure*}

In \cref{fig:shoulder-actions-comp}, we see the distribution of the number of actions needed to complete a scenario successfully for the three different action selection methods. Across both failure cases and all scenarios, we see a trend of the Greedy ASM requiring, on average, the fewest actions to complete the given task. This strongly supports the usefulness of including a simulator in planning to inform the best action. When comparing the Lazy and Random ASMs, we see that the Random approach usually has a flatter or wider multi-peak distribution. This is expected as the Random approach can complete the task in one action or require many actions. The Lazy ASM is a balance between the two extremes. In Failure Case 1, the average number of actions sits between the Greedy and Random ASMs. In contrast, in Failure Case 2, we see the weakness of its one-shot approach, taking one to two more actions, on average, to complete than the Random or Greedy approach. Still, there is less deviation than the Random approach, with a shorter tail on the higher end of the actions needed.

Overall, our data shows a new and unique ability to conduct manipulation tasks while experiencing LMJs that restrict the robot's capability to manipulate using traditional pick-and-place approaches. Our sim-in-the-loop planner improved the robot's ability to conduct these tasks by looping in a physics model to our NPM actions. We demonstrated that a balanced approach of speed and simulation accuracy can complete our manipulation scenarios well. We have \textbf{shown that while experiencing LMJs, our approach can unlock the capability of robotic manipulators to conduct tabletop manipulation successfully, supporting H.2.}

\section{Conclusion and Discussion}\label{sec:conclusion}
Our research has demonstrated the power of non-prehensile manipulation and whole-body interaction in enabling robotic manipulators to operate effectively despite locked multi-joint failures. By leveraging these two key strategies, we have shown that our approach can significantly increase the manipulable area, allowing the robot to maintain up to 92\% of the nominal reachable area, even when the gripper is unusable. Furthermore, our sim-in-the-loop planner effectively utilizes kinodynamic maps to complete manipulation tasks during LMJ failures. 

We believe future work can: i) explore and improve the completeness of our motion planner \cite{kleinbort2018probabilistic}; ii) expand the portfolio of motion primitives, such as pushing and rolling; iii) extend our approach to more precise manipulation tasks and other robot embodiments, like mobile manipulators; and iv) improve the simulation physics to match the real world more closely. Furthermore, we believe that applying optimization-based or learned approaches to joint failures can be fruitful areas of future research. This research paves the way for robots to operate independently for extended periods, even in the face of significant failures.

\AtNextBibliography{\small}
\printbibliography

\end{document}